\documentclass[a4paper,twoside]{article}
\usepackage[pdftex,
            pdfauthor={Andrea Asperti and Claudio Mastronardo},
            pdftitle={The Effectiveness of Data Augmentation for Detection of Gastrointestinal Diseases
from Endoscopical Images},
            pdfsubject={Application of data augmentation, deep learning and ensembling to the gastrointestinal disease detection from images},
            pdfkeywords={Data Augmentation, Deep Learning, Gastrointestinal Disease, Endoscopy, Kvasir},
            pdfproducer={Latex with hyperref},
            pdfcreator={pdflatex}]{hyperref}
\usepackage{amsmath}
\usepackage{epsfig}
\usepackage{subfigure}
\usepackage{calc}
\usepackage{amssymb}
\usepackage{amstext}
 \usepackage{multirow}
\usepackage{amsmath}
\usepackage{bm}
\usepackage{amsthm}
\usepackage{multicol}
\usepackage{pslatex}
\usepackage{apalike}
\usepackage{hyperref}
\usepackage{mathtools}
\DeclarePairedDelimiter\ceil{\lceil}{\rceil}

\usepackage[utf8]{inputenc}
\usepackage{SCITEPRESS}     

\begin{document}

\title{The Effectiveness of Data Augmentation for Detection of Gastrointestinal Diseases
from Endoscopical Images}

\author{\authorname{Andrea Asperti \sup{1} and  Claudio Mastronardo\sup{1}}
\affiliation{\sup{1}Department of Informatics: Science and Engineering (DISI), University of Bologna, Mura Anteo Zamboni 7, 40127, Bologna, Italy}
\email{ andrea.asperti@unibo.it, claudio.mastronardo@studio.unibo.it}
}


\abstract{The lack, due to privacy concerns, of large public databases of medical pathologies is a well-known and major problem, substantially hindering the application of deep learning techniques in this field. In this article, we investigate the possibility to supply to the deficiency in the number of data by means of data augmentation techniques, working on the recent
$Kvasir$ dataset \cite{Pogorelov:2017:KMI:3083187.3083212} of endoscopical
images of gastrointestinal diseases. The dataset comprises 4,000 colored images labeled and verified by medical endoscopists, covering a few common pathologies at different anatomical landmarks: Z-line, pylorus and cecum.
We show how the application of data augmentation techniques allows to
achieve sensible improvements of the classification with respect to previous approaches, both in terms of precision and recall.}

\onecolumn \maketitle \normalsize \vfill

\section{\uppercase{Introduction}}
\label{sec:introduction}
Gastrointestinal diseases affect 60 to 70 million of people every year in the United States \cite{NIDDK}.
Diagnosis of such diseases has to be done by a trained gastroenterologist. Such diagnosis often involves one or more invasive and not invasive endoscopic examinations enabling a direct and visual feedback of the status of internal organs. In this case, it is essential to be able to perform a detailed image analysis in order to diagnose the disease. For example, the
degree of inflammation directly affects the choice of therapy in inflammatory bowel diseases (IBD) \cite{doi:10.1016/j.crohns.2013.09.010}.
In recent years, automatic elaboration of digital images has seen an enormous increment of research interest due to latest impressive results on many computer vision sub-related tasks. Such results almost always involved deep learning based algorithms.
Notoriously, deep learning techniques frequently require a very large amount of training examples, and the availability of several such large datasets\cite{imagenet_cvpr09}\cite{cifar10} has heavily contributed to the evolution of the field. To make an example, ImageNet is composed of over
14 million images, spread over 22K different categories.
 
Since automatic detection, recognition and assessment of pathological findings can provide a valid assistance for doctors in their diagnosis, there is a growing demand for medical datasets, especially in relation with the application of deep learning techniques in this field. 

A recent example of such a dataset for gastrointestinal diseases is $Kvasir$ \cite{Pogorelov:2017:KMI:3083187.3083212}, comprising about
4,000 colored images labeled and verified by medical endoscopists 
(for details on the dataset and the pathologies see 
Section~\ref{Sec:dataset}).

Unfortunately, the dataset is quite small for the purposes of deep
learning. This is a well-known problem of this field: building large 
databases of labeled information is not only an 
expensive operation, requiring the supervision of an expert, but in
the case of medical pathologies, it is even more difficult due
to the privacy constraints preventing the publication of sensible data.

In this article, following similar successful attempts made on different datasets (see Section~\ref{Sec:Related}), we show that 
{\em data augmentation} can provide a valid palliative to the small
dimension of the above mentioned dataset, 
proving that the problem of automatic diagnosing of gastrointestinal diseases from images can be successfully addressed by means of deep learning algorithms. Specifically we make use of transfer learning \cite{pmlr-v27-bengio12a}, Convolutional Neural Networks (CNNs) \cite{doi:10.1162/neco.1989.1.4.541}, data augmentation techniques (see e.g. \cite{DBLP:journals/corr/WongGSM16} for a recent survey) and snapshot ensembling \cite{DBLP:journals/corr/HuangLPLHW17}, obtaining
sensible improvements in the classification with respect to previous approaches, both in terms of precision and recall.

The structure of the article is the following. In 
Section~\ref{Sec:Related} we discuss related works, 
especially from the point of view of data augmentation.
Section~\ref{Sec:dataset} contains a detailed description 
of the Kvasir dataset, used for our experiments.
In Section~\ref{Sec:approach}, we explain our methodology.
The experimental results are reported in Section~\ref{Sec:results}. Section~\ref{Sec:future} is devoted
to our plans for future research on this topic. Finally, 
a few concluding remarks are given in Section~\ref{Sec:conclusions}.


\section{\uppercase{Related work}}\label{Sec:Related}
Data augmentation is a key technique of machine learning. It consists
in increasing the number of data, by artificially synthesizing new
samples from existing ones, usually via minor perturbations. For
instance, in the case of images, typical operations are rotation,
lighting modifications, rescaling, cropping and so on; even adding
random noise can be seen as a form of data augmentation.
Usually deployed as a
means for reducing overfitting and improving the robustness of systems
(see e.g. \cite{DBLP:conf/aist/PrisyachMU16} for a recent application to
sound recognition), it frequently proved to be 
also useful for improving the performance of deep learning techniques,
especially in presence of a low number of training data. 
In the field of image processing, a sophisticated form of data augmentation (the so called {\em fancy} PCA technique) was a key ingredient of the famous AlexNet \cite{NIPS2012_4824}. 
More recently, massive data augmentation was exploited in \cite{Farfade15}, where for the first time a {\em single} deep architectural
network was trained to detect faces under unconstrained conditions, and
in a wide range of different orientations. Similarly, addressing a problem
of relational classification in Natural Language Processing, \cite{DBLP:journals/corr/XuJMLCLJ16} have been able to outperform previous shallow neural nets by just augmenting the number of input sentences by
means of simple grammatical manipulations.
In the field of medicine, data augmentation has been 
very recently applied in \cite{DBLP:journals/corr/VasconcelosV17} in relation with the ISBI 2017 Melanoma Classification Challenge (named Skin Lesion Analysis towards Melanoma Detection), successfully overcoming the small dimension and biased nature of the biological database.

A large number of different augmentation techniques has been recently
compared in \cite{augmentationWang}, comprising sophisticated techniques
based on Generative Adversarial Networks \cite{GAN}, using the CycleGan tool \cite{DBLP:journals/corr/ZhuPIE17}. According to this study, traditional
augmentation techniques remain the most successful, motivating our choice of sticking to them in this work.

\section{\uppercase{Dataset}}\label{Sec:dataset}
For our experiments, we worked on the recently published $Kvasir$ dataset \cite{Pogorelov:2017:KMI:3083187.3083212}. The Kvasir dataset has been created in order to be used to improve applications involving automatic detection, classification and localization of endoscopic pathological findings in images captured in the gastrointestinal tract. This new dataset comprises of 4,000 colored images\footnote{We used the first version of the dataset. In date 17/10/2017 a second version of the Kvasir dataset has been released. This new version has 8,000 images.} labeled and verified by medical endoscopists. It has 8 classes representing several diseases as well as normal anatomical landmarks. The dataset has 500 examples for each class, making it perfectly balanced.

\begin{figure}

\centering     
\subfigure[Ulcerative colitis]{\label{fig:a}\includegraphics[width=37mm]{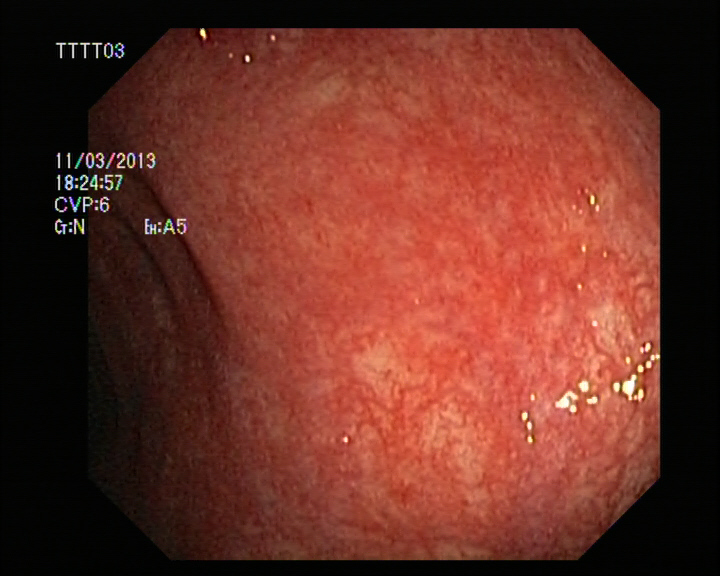}} \subfigure[Dyed lifted polyp]{\label{fig:b}\includegraphics[width=37mm]{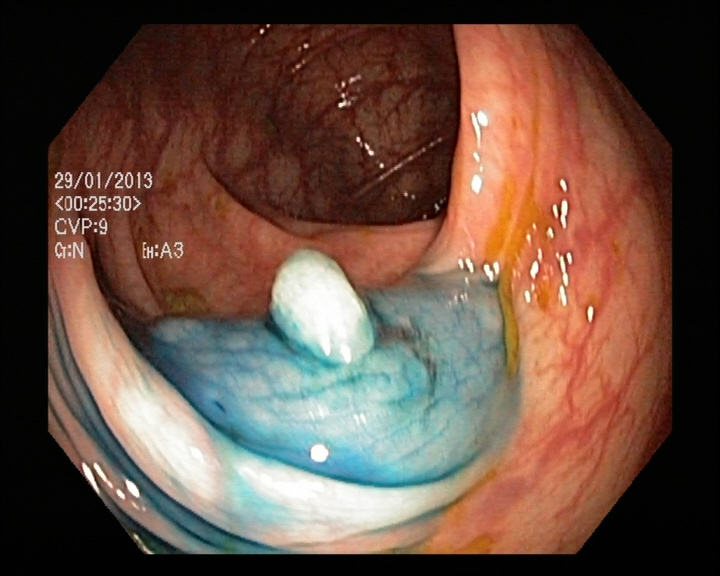}}
\subfigure[Dyed resection margin]{\label{fig:c}\includegraphics[width=37mm]{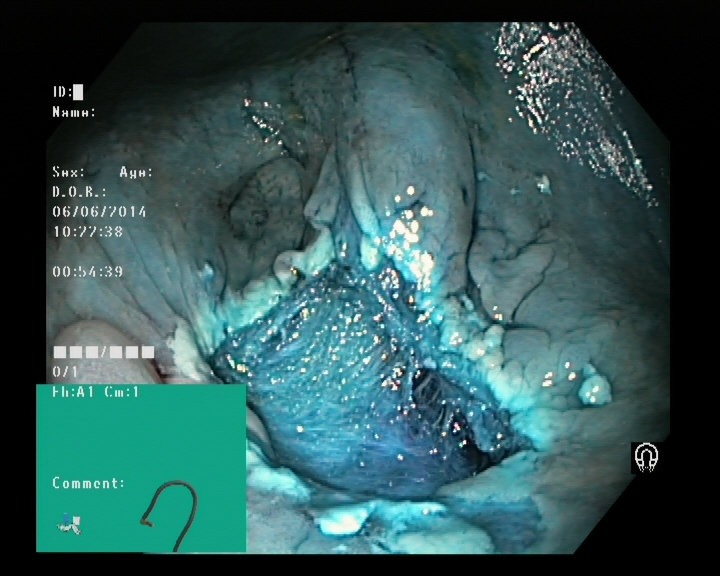}} \subfigure[Normal z-line]{\label{fig:d}\includegraphics[width=37mm]{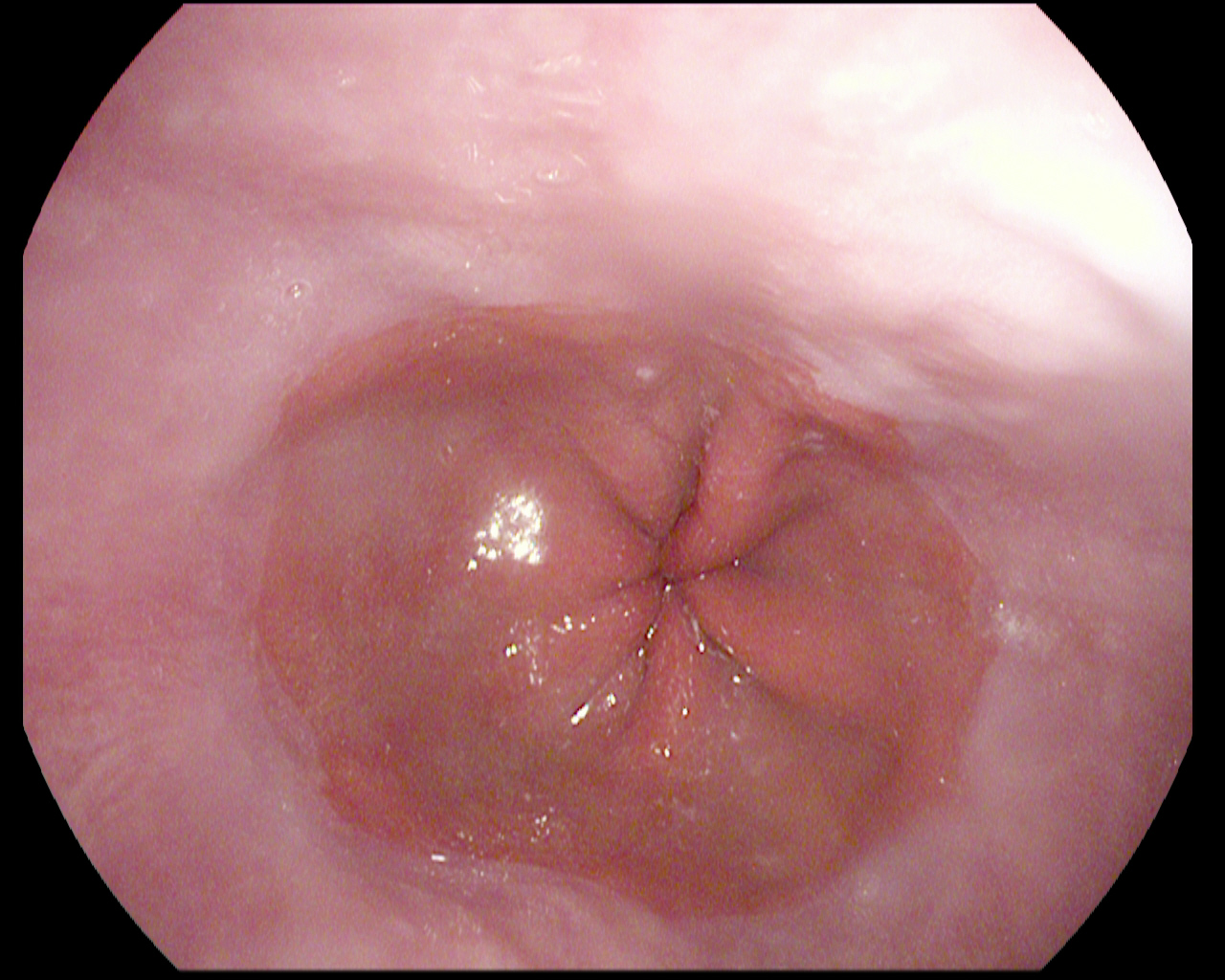}}

\caption{Some images extracted from the KVASIR dataset}
\label{fig1}
\end{figure}

The anatomical landmarks are: Z-line, pylorus and cecum. Diseases: esophagitis, polyps and ulcerative colitis. There are also images representing dyed and lifted polyps and dyed resection margins. Images across the dataset have resolution from 720x576 up to 1920x1072 pixels. Some extracted images are shown in Figure~\ref{fig1}.
\section{Approach}\label{Sec:approach}
Our approach is an ensemble of models created by using transfer learning from 
previously trained convolutional neural nets and data
augmentation.

\subsection{Transfer learning}
In order to save computation time and focus on the high level representations learned by CNNs we used a transfer learning approach\cite{pmlr-v27-bengio12a}. We used Inception v3 model\cite{DBLP:conf/cvpr/SzegedyVISW16} and Keras library\cite{Keras} with Tensorflow\cite{Tensorflow2015} as backend. We loaded pre-trained weights learned on the Imagenet\cite{imagenet_cvpr09} dataset and cut the last dense layers. After the last convolutional layer we added a global averaging pooling layer, a dense layer with 1024 neurons with ReLU\cite{icml2010_NairH10} as activation function and finally a softmax layer of 8 neurons, one for every class.
All images have been resized to a resolution of 299x299 in order to be fed to Inception v3.

We froze all Inception's already trained layers and used Adam optimizer\cite{journals/corr/KingmaB14} to tune last dense layers' weights. Categorical cross-entropy has been used as the loss function.

After several epochs we started modifying both last dense layers' weights and convolutional layers from the top 2 inception blocks from Inception v3. We switched to stochastic gradient descent\cite{Zhang:2004:SLS:1015330.1015332} with momentum, enabling us to use a very small learning rate (0.0001) in order to make sure that the magnitude of the updates stays very small and does not break previously learned features. We trained for about 17 epochs (losses for the fine tuning phase in Figure \ref{losses}). In both fine-tuning phases a batch size of 16 instances has been used.
\subsection{Data augmentation}
A key role in our results has been represented by using several data augmentation techniques. In order to make our model more robust, prevent overfitting and enabling it to generalize better we used Keras' utilities to augment training instances by applying several random transformations. Values for parameters' based transformations have been picked randomly in defined ranges.  A list of data augmentation transformations (and their chosen range of action) used during training is reported in the table \ref{table_label}.
\begin{table}
\centering
\caption{Data augmentation transformations and their range values.}
\label{table_label}
\begin{tabular}{ll}
\hline
\textbf{Type}         & \textbf{Range}           \\ \hline
Rotation     & {[}-30 $^{\circ}$, +30 $^{\circ}${]} \\
Width shift  & 0.1             \\
Height shift & 0.1             \\
Shear        & 0.2             \\
Zoom         & {[}0.8, 1.1{]}   \\ \hline  
\end{tabular}
\end{table}

Since images were black bordered we didn't use much of zooming out to prevent the generation of images having too much black component. When having to fill pixels due to zooming out and shifting we adopted a nearest pixel policy, repeating nearest pixel value across the axis.
Moreover we used random horizontal flips and vertical flips.

To normalize both training and test data we divided every pixel's color value by 255 in order to have all pixel values in the range $[0,1]$.

\begin{figure}

\centering     
\subfigure[Original image]{\label{fig:a2}\includegraphics[width=37mm]{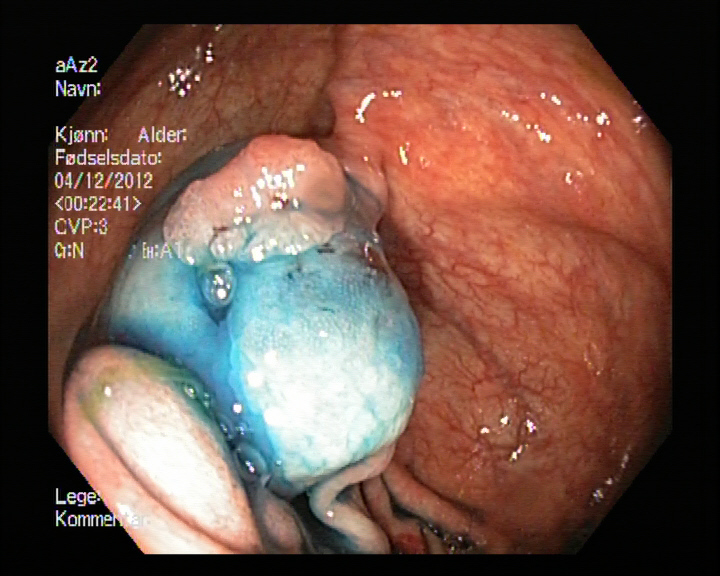}} \subfigure[Shearing and rotation]{\label{fig:b2}\includegraphics[width=37mm]{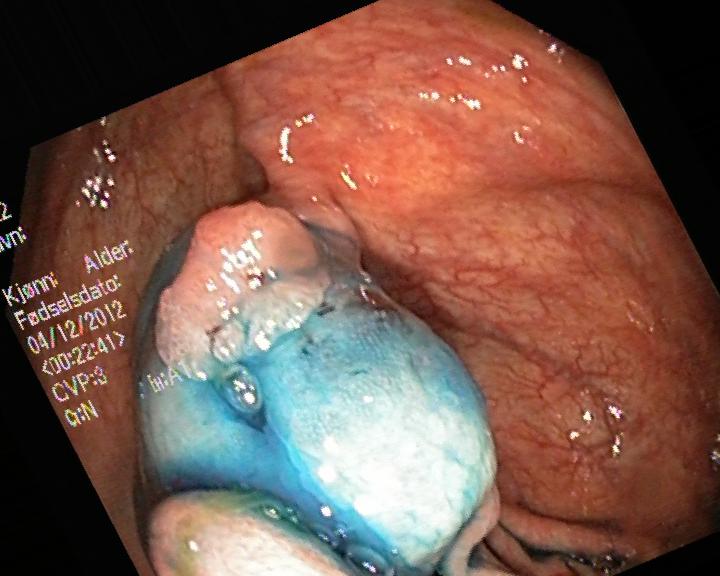}}
\subfigure[Rotation and shifting]{\label{fig:c2}\includegraphics[width=37mm]{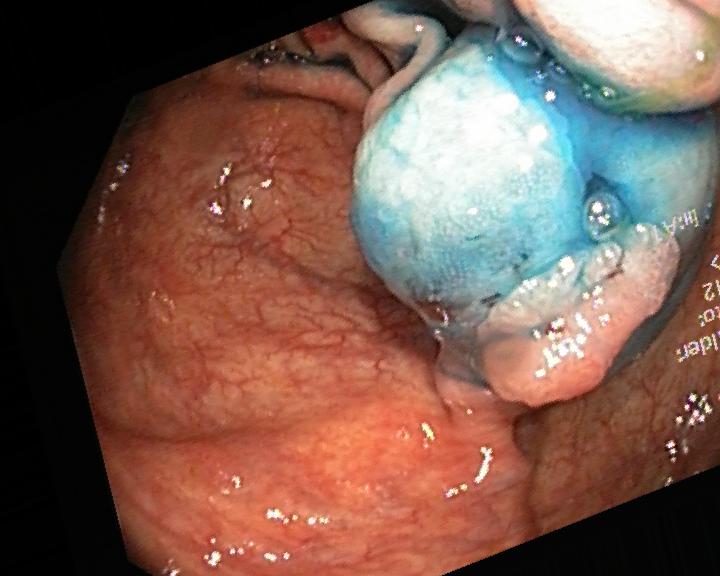}} \subfigure[Rotation and zooming]{\label{fig:d2}\includegraphics[width=37mm]{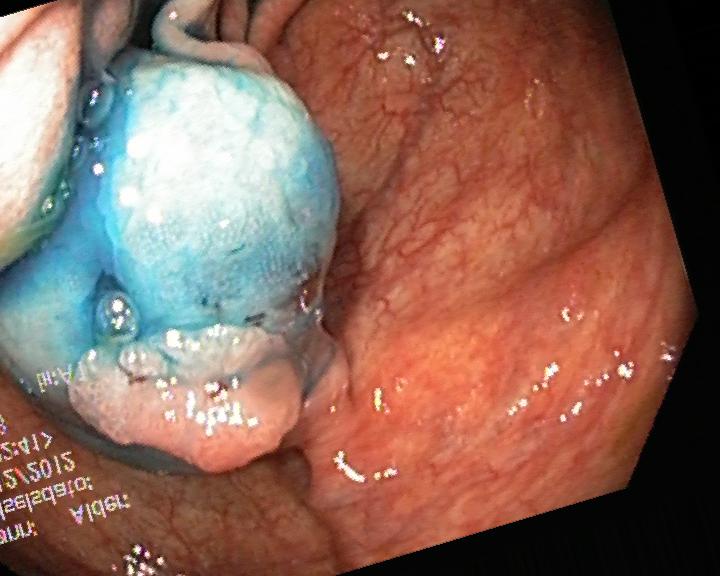}}

\caption{Augmented examples}
\label{data_aug}
\end{figure}

During training we kept generating new images following this data augmentation policy, never feeding the same images to the network. Some examples of augmented images are reported in figure \ref{data_aug}.
\subsection{Snapshot Ensembling}
\begin{figure*}
\centering
\includegraphics[width=\linewidth]{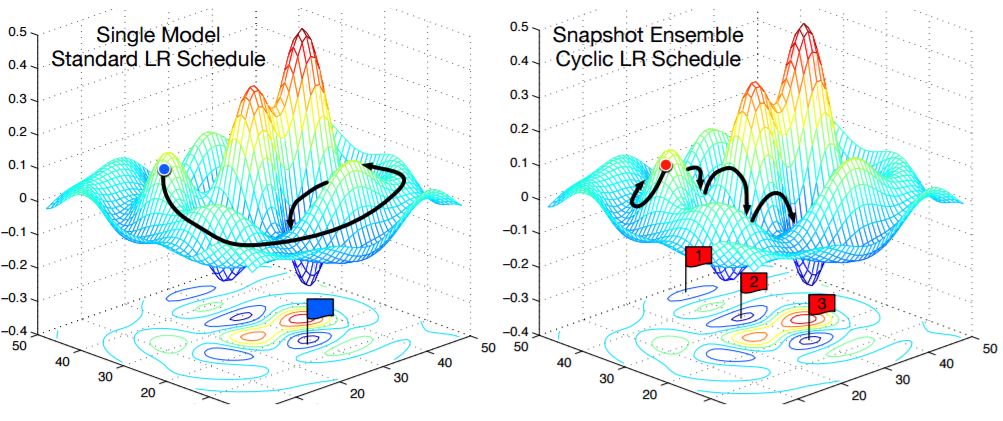}
\caption{\textbf{Left}: Classic SGD. \textbf{Right}: Snapshot ensembling converging to several minima and taking snaphots. Image borrowed from \cite{DBLP:journals/corr/HuangLPLHW17}}
\label{snapshot_image}
\end{figure*}

To improve classification precision and avoid to be trapped in local minima, we adopted an ensembling approach. In particular, we used Snapshot ensembling \cite{DBLP:journals/corr/HuangLPLHW17} allowing us to execute one training but getting several models. Snapshot Ensembling is a method to obtain multiple neural networks at no additional training cost. This is achieved by letting a single model converge into several different local minima along its optimization path on the error surface. Saving network weights at certain epochs constitutes saving several ''snapshots'' (see Figure \ref{snapshot_image} for a visual representation).
Since, in general, there exist multiple local minima, snapshot ensembling let's the current model dive into a minima using a decreasingly learning rate value, save the snapshot at that minimum and then increase the learning rate in order to escape the local minima and attempt to find another possibly better minima. This repeated rapid convergence is achieved taking advantage of cosine annealing cycles as the learning rate schedule. The learning rate is achieved by :

\centerline{
$\alpha(t)\ =\ \frac{\alpha_0}{2}\Big(\cos\Big(\frac{\pi\  mod(t-1,\ceil*{T/M}}{\ceil*{T/M}}\Big)+1\Big)$\\[2mm]
}
where $\alpha_0$ stands for the initial learning rate, $t$ is the current epoch, $T$ is the total number of epochs and $M$ is the chosen number of models in the ensemble. For our experiments we used an initial learning rate of 0.1, we trained for about 22 epochs and we've chosen an ensemble with 5 models ($T=5$).

\section{Experimental results}\label{Sec:results} 

\subsection{Classification metrics}
Following \cite{Pogorelov:2017:KMI:3083187.3083212}, classification has been tested using
traditional metrics like precision, recall, F1 score
and accuracy. Precision is the fraction of relevant instances (True Positives) among the retrieved instances, while recall (or sensitivity) is the fraction of relevant instances that have been retrieved over the total amount of relevant instances; 
F1-score is a simple combination of precision and recall expressed in terms
of their harmonic mean; finally, accuracy is simply the fraction of correctly 
classified samples.

While the notions of precision and recall are clear in the case of a binary classification problem,
their generalization to multiclass classification is not entirely 
straightforward. There are several possible ways to combine
results across labels, and unfortunately \cite{Pogorelov:2017:KMI:3083187.3083212} 
are not explicit about the method they used. For this reason, we tested
several of them, whose precise definition is given below.
Fortunately, results are very similar, and we shall only 
report them for the so called "micro" averaging.

Let us introduce the following notation

\begin{itemize}
\item let $y$ be the set of {\em predicted} (input, label) pairs
\item let $\hat{y}$ be the set of {\em true} (input, label) pairs
\item let $L$ be the set of labels
\item let $S$ be the set of samples
\item let $y_s$ ($\hat{y}_s$) be the subset of $y$ (resp. $\hat{y})$ with sample $s$
\item let $y_l$ ($\hat{y}_l$) be the subset of $y$ (resp. $\hat{y}$) with label $l$
\item let $P(A,B) = \frac{|A\cap B|}{A}$
\item let $R(A,B) = \frac{|A \cap B|}{B}$
\item let $F_1(A,B) = \frac{P(A,B)\times R(A,B)}{P(A,B)+R(A,B)}$
\end{itemize}

In Figure~\ref{av}, we give the formal definition of the most typical forms of averaging.

\begin{figure*}[t]

\begin{center}
\begin{tabular}{lccc}
{\bf Average} & {\bf Precision} & {\bf Recall} & {\bf \bm{$F_1$}} \\
micro & $P(y,\hat{y})$ & $R(y,\hat{y})$ & $F_1(y,\hat{y})$\\
samples & $\frac{1}{|S|}\sum_{s \in S}P(y_s,\hat{y}_s)$ & $\frac{1}{|S|}\sum_{s \in S}R(y_s,\hat{y}_s)$ & $\frac{1}{|S|}\sum_{s \in S}F_1(y_s,\hat{y_s})$\\
macro & $\frac{1}{|L|}\sum_{l \in L}P(y_l,\hat{y}_l)$ & $\frac{1}{|L|}\sum_{l \in L}R(y_l,\hat{y}_l)$ & $\frac{1}{|L|}\sum_{l \in L}F_1(y_l,\hat{y_l})$\\
weighted & $\frac{1}{\sum_{l \in L}|\hat{y}_l|}\sum_{l \in L}|\hat{y}_l|P(y_l,\hat{y}_l)$ & 
$\frac{1}{\sum_{l \in L}|\hat{y}_l|}\sum_{l \in L}|\hat{y}_l|R(y_l,\hat{y}_l)$ & 
$\frac{1}{\sum_{l \in L}|\hat{y}_l|}\sum_{l \in L}|\hat{y}_l|F_1(y_l,\hat{y_l})$
\end{tabular}

\caption{\label{av}Typical averaging techniques for classification metrics.}
\end{center}

\end{figure*}

\subsection{Evaluation}
\begin{table}[]
\centering
\caption{Confusion matrix produced by the ensemble. A=Dyed lifted polyps, B=Dyed resection margins, C=Esophagitis, D=Normal cecum, E=Normal pylorus, F=Normal z-line, G=Polyps and H=Ulcerative colitis.}
\label{confusion_lab}
\begin{tabular}{llllllllll}
                                                      & \multicolumn{8}{c}{\large Actual class}                                                                                                                                                                             &                         \\ 
{\multirow{9}{*}{\rotatebox{90}{\large Predicted class}}} & 
\multicolumn{1}{l|}{}
& \multicolumn{1}{l|}{\textbf{A}}  & \multicolumn{1}{l|}{\textbf{B}}  & \multicolumn{1}{l|}{\textbf{C}}  & \multicolumn{1}{l|}{\textbf{D}}  & \multicolumn{1}{l|}{\textbf{E}}  & \multicolumn{1}{l|}{\textbf{F}}  & \multicolumn{1}{l|}{\textbf{G}}  & \multicolumn{1}{l}{\textbf{H}}  \\ \cline{2-10} 
& \multicolumn{1}{l|}{\textbf{A}} & \multicolumn{1}{l|}{46} & \multicolumn{1}{l|}{8}  & \multicolumn{1}{l|}{0}  & \multicolumn{1}{l|}{0}  & \multicolumn{1}{l|}{0}  & \multicolumn{1}{l|}{0}  & \multicolumn{1}{l|}{0}  & \multicolumn{1}{l}{0}  \\ \cline{2-10} 
& \multicolumn{1}{l|}{\textbf{B}} & \multicolumn{1}{l|}{4}  & \multicolumn{1}{l|}{42} & \multicolumn{1}{l|}{0}  & \multicolumn{1}{l|}{0}  & \multicolumn{1}{l|}{0}  & \multicolumn{1}{l|}{0}  & \multicolumn{1}{l|}{0}  & \multicolumn{1}{l}{0}  \\ \cline{2-10} 
& \multicolumn{1}{l|}{\textbf{C}} & \multicolumn{1}{l|}{0}  & \multicolumn{1}{l|}{0}  & \multicolumn{1}{l|}{39} & \multicolumn{1}{l|}{0}  & \multicolumn{1}{l|}{0}  & \multicolumn{1}{l|}{7}  & \multicolumn{1}{l|}{0}  & \multicolumn{1}{l}{0}  \\ \cline{2-10} 
& \multicolumn{1}{l|}{\textbf{D}} & \multicolumn{1}{l|}{0}  & \multicolumn{1}{l|}{0}  & \multicolumn{1}{l|}{0}  & \multicolumn{1}{l|}{50} & \multicolumn{1}{l|}{0}  & \multicolumn{1}{l|}{0}  & \multicolumn{1}{l|}{1}  & \multicolumn{1}{l}{0}  \\ \cline{2-10} 
& \multicolumn{1}{l|}{\textbf{E}} & \multicolumn{1}{l|}{0}  & \multicolumn{1}{l|}{0}  & \multicolumn{1}{l|}{0}  & \multicolumn{1}{l|}{0}  & \multicolumn{1}{l|}{50} & \multicolumn{1}{l|}{0}  & \multicolumn{1}{l|}{1}  & \multicolumn{1}{l}{0}  \\ \cline{2-10} 
& \multicolumn{1}{l|}{\textbf{F}} & \multicolumn{1}{l|}{0}  & \multicolumn{1}{l|}{0}  & \multicolumn{1}{l|}{11} & \multicolumn{1}{l|}{0}  & \multicolumn{1}{l|}{0}  & \multicolumn{1}{l|}{43} & \multicolumn{1}{l|}{0}  & \multicolumn{1}{l}{0}  \\ \cline{2-10} 
& \multicolumn{1}{l|}{\textbf{G}} & \multicolumn{1}{l|}{0}  & \multicolumn{1}{l|}{0}  & \multicolumn{1}{l|}{0}  & \multicolumn{1}{l|}{0}  & \multicolumn{1}{l|}{0}  & \multicolumn{1}{l|}{0}  & \multicolumn{1}{l|}{47} & \multicolumn{1}{l}{1}  \\ \cline{2-10} 
& \multicolumn{1}{l|}{\textbf{H}} & \multicolumn{1}{l|}{0}  & \multicolumn{1}{l|}{0}  & \multicolumn{1}{l|}{0}  & \multicolumn{1}{l|}{0}  & \multicolumn{1}{l|}{0}  & \multicolumn{1}{l|}{0}  & \multicolumn{1}{l|}{1}  & \multicolumn{1}{l}{49} \\ 

\end{tabular}

\end{table}
\begin{figure}

\centering     
\subfigure[Training loss]{\label{fig:a3}\includegraphics[width=38mm]{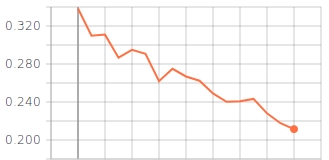}}\subfigure[Test loss]{\label{fig:b3}\includegraphics[width=38mm]{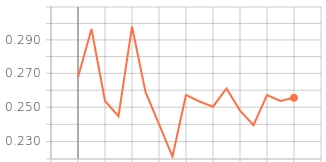}}
\caption{Categorical cross-entropy error in function of training and test epochs}
\label{losses}
\end{figure}

We computed the metrics from the produced confusion matrix (see \ref{confusion_lab}), in order to compare our approach to the previous ones \cite{Pogorelov:2017:KMI:3083187.3083212} splitting the dataset into training and test sets.\\ Results are reported in table \ref{results}.
All metrics have been computed using the \verb+precision_recall_fscore_support+ function of scikit-learn 
\cite{scikit-learn}.

\begin{table}
\centering
\caption{Our metrics compared to the best ones reported in \cite{Pogorelov:2017:KMI:3083187.3083212}. All metrics are micro averaged.}
\label{results}
\resizebox{\linewidth}{!}{
\begin{tabular}{llllll}
\hline
\multicolumn{1}{|l|}{\textbf{Method}}   & \multicolumn{1}{l|}{\textbf{PREC}} & \multicolumn{1}{l|}{\textbf{REC}} & \multicolumn{1}{l|}{\textbf{ACC}} & \multicolumn{1}{l|}{\textbf{F1}} & \multicolumn{1}{l|}{\textbf{MCC}} \\ \hline
2 GF Logistic Model Tree                & 0.706                              & 0.707                             & 0.926                             & 0.705                            & 0.664                               \\
6 GF Random Forest                      & 0.732                              & 0.732                             & 0.933                             & 0.727                            & 0.692                               \\
6 GF Logistic Model Tree                & 0.748                              & 0.748                             & 0.937                             & 0.747                            & 0.711                                \\
\textbf{Ensemble of Inception+}\\\textbf{fine tuning+}\\\textbf{data augmentation} & \textbf{0.915}                    & \textbf{0.915}                   & \textbf{0.915}                   & \textbf{0.915}                  &              \textbf{0.903}                    
\end{tabular}
}
\end{table}

Our model achieves better scores for precision, recall and f-measure 
while essentially preserving the same accuracy with respect to the previous tested solutions\cite{Pogorelov:2017:KMI:3083187.3083212}. We found that the model is particularly precise in classifying examples belonging to normal cecum and normal pylorus.

\begin{figure}

\centering     
\subfigure[predicted: lifted polyp\newline actual: resection margin]{\label{fig:aa}\includegraphics[width=48mm]{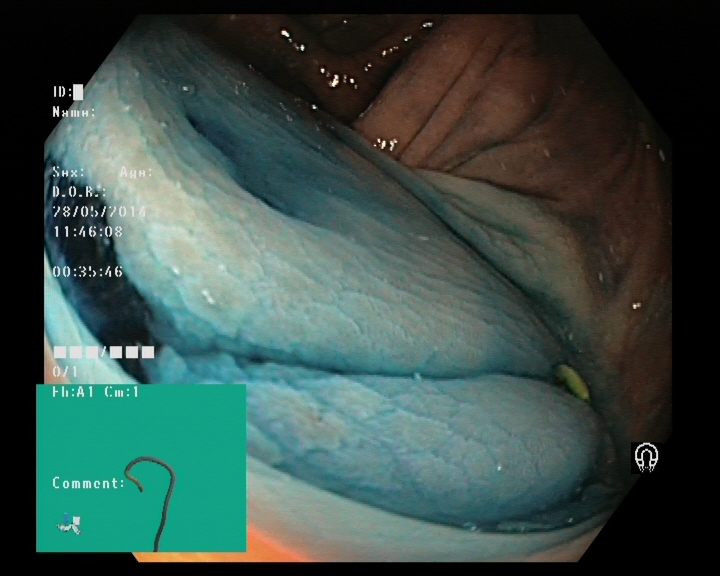}}\\
\subfigure[predicted: resection margin\newline actual: lifted polyp]{\label{fig:bb}\includegraphics[width=48mm]{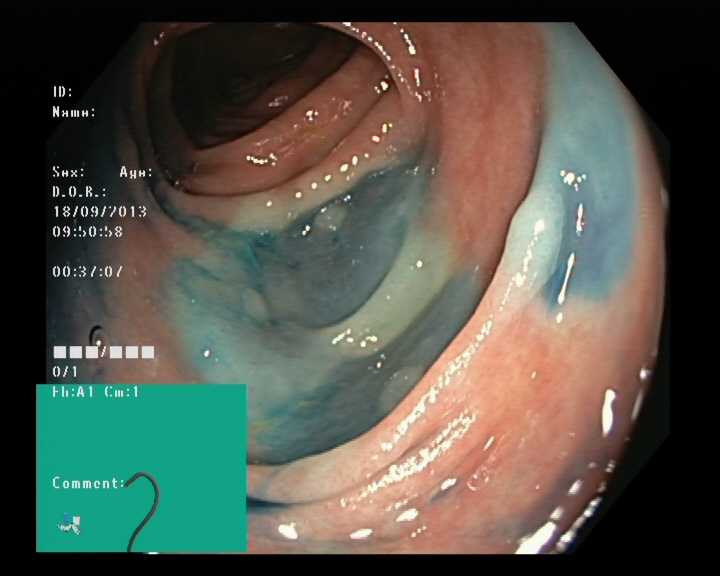}}\\
\subfigure[predicted: esophagitis\newline actual: normal z-line]{\label{fig:cc}\includegraphics[width=48mm]{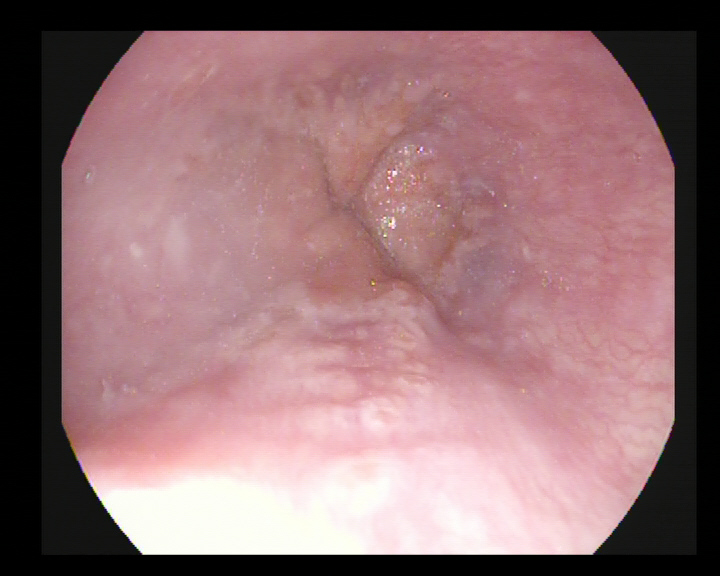}}
\caption{Some misclassified samples}
\label{misclassified}
\end{figure}

Misclassifications mostly involve dyed lifted polyps and dyed resection margins (e.g. see figure \ref{misclassified} for some examples). In fact, these two classes are made up of very similar images, having the same amount of blue color. Moreover some other misclassified instances belong to normal z-line and esophagitis. This is reasonable since some cases of esophagitis are not so clearly spotted in images, where
it may be confused with the gastroesophageal junction that joins the esophagus to the stomach. An example is reported in figure \ref{misclassified} (c) where the classifier predicted esophagitis instead of z-line. This error might be related to  specific z-line tissues being visually similar to an esophagitis of grade A \cite{Lundell172} (lowest inflammatory grade).

Misclassifications could be possibly overcame trying to train the network for a greater number of epochs, or working with the new extended version of the dataset. Prediction confusion might be improved increasing the number of samples from the dyed lifted polyps and dyed resection margins as well as from z-line and esophagitis classes.

\section{Future work}\label{Sec:future}
Several deep convolutional neural networks have been published since Inception v3, such as \cite{huang2017densely}, \cite{He2015} \cite{DBLP:journals/corr/ZhuPIE17}, \cite{DBLP:journals/corr/WongGSM16}, \cite{DBLP:journals/corr/XuJMLCLJ16}. Experiments can be done using these newly proposed architectures in conjunction with data augmentation techniques.


Stacking additional dense layers can be another direction worth to be investigated, as well as making a more exhaustive experimentation with different activation functions such ELU \cite{DBLP:journals/corr/ClevertUH15}, LeakyRelu \cite{DBLP:journals/corr/ZhuPIE17}, Swish \cite{DBLP:journals/corr/abs-1710-05941} etc.

A different investigation might consist in visualizing high level learned features from the last convolutional layers, in order to improve our
grasp of the discriminative characteristics learned by the network.

All our experiments have been conducted over the first version of the Kvasir dataset;
repeting training and validation on the recently released extended version would 
provide an important additional validation of our methodology.

Finally, it would be particularly useful to further extend the Kvasir dataset with new classes, in order to meet diagnosis needs in the direction of several other very known and diffused diseases such as Chron's disease. We are currently exploring the
possibility to cooperate with the gastroenterology department of the Sant'Orsola Hospital in Bologna to extend the dataset along these lines.

\section{\uppercase{Conclusions}}\label{Sec:conclusions}
\label{sec:conclusion}
In this work we addressed the problem of gastrointestinal disease detection and identification. By a simple combination of Convolutional Neural Networks, transfer learning, and data augmentation we outperfomed previous techniques in terms of precision, recall, and f-measure, while essentially preserving the same accuracy. Our experimentation confirms once more that data augmentation is a viable technique for boosting deep learning in presence of small dataset.

\vfill
\bibliographystyle{apalike}
{\small
\bibliography{example}}

\vfill
\end{document}